%% file: main.tex
\documentclass[letterpaper, 10 pt, conference]{ieeeconf}
\IEEEoverridecommandlockouts
\input{preamble}

\input{acronyms}

\title{\LARGE \bf 
Grasping in Uncertain Environments: A Case Study For Industrial Robotic Recycling}

\author{Annalena Daniels$^{1}$ and Sebastian Kerz$^{1}$ and Salman Bari$^{1}$ and Volker Gabler$^{1}$ and Dirk Wollherr$^{1}$
\thanks{© 2023 IEEE. Personal use of this material is permitted.  Permission from IEEE must be obtained for all other uses, in any current or future media, including reprinting/republishing this material for advertising or promotional purposes, creating new collective works, for resale or redistribution to servers or lists, or reuse of any copyrighted component of this work in other works. This work was accepted and presented at the IEEE International Conference on Systems, Man and Cybernetics. DOI: 10.1109/SMC53992.2023.10394008}
\thanks{*The research leading to these results has received funding from the Horizon 2020 research and innovation programme under grant agreement No. 820742 of the project ”HR-Recycler - Hybrid Human-Robot RECYcling plant for electriCal and eLEctRonic equipment”. }
\thanks{$^{1}$ All authors are with the Chair of Automatic Control Engineering, Technical University of Munich, Theresienstrasse 90, 80333 Munich, Germany. 
        {\tt\small {a.daniels, s.kerz, s.bari, volker.gabler, dw}@tum.de}}%
}

\begin{document}

\maketitle
\thispagestyle{empty}
\pagestyle{empty}

\begin{abstract} 
Autonomous robotic grasping of uncertain objects in uncertain environments is an impactful open challenge for the industries of the future. One such industry is the recycling of \ac{WEEE} materials, in which electric devices are disassembled and readied for the recovery of raw materials. Since devices may contain hazardous materials and their disassembly involves heavy manual labor, robotic disassembly is a promising venue. However, since devices may be damaged, dirty and unidentified, robotic disassembly is challenging since object models are unavailable or cannot be relied upon. This case study explores grasping strategies for industrial robotic disassembly of \ac{WEEE} devices with uncertain vision data. We propose three grippers and appropriate tactile strategies for force-based manipulation that improves grasping robustness. For each proposed gripper, we develop corresponding strategies that can perform effectively in different grasping tasks and leverage the grippers design and unique strengths. Through experiments conducted in lab and factory settings for four different \ac{WEEE} devices, we demonstrate how object uncertainty may be overcome by tactile sensing and compliant techniques, significantly increasing grasping success rates. 
\end{abstract}
\vspace{-0.2cm}
\begin{figure}[htb]
    \centering
    \includegraphics[width = 0.37 \textwidth]{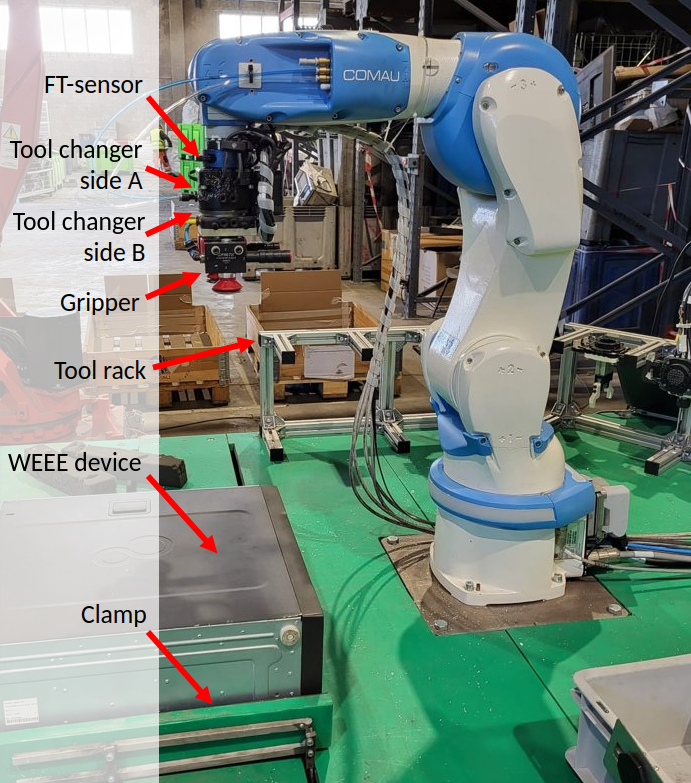}
    \caption{Industrial 6-DoF robot arm in industrial environment with \ac{FT} sensor and gripper attached through tool changer. A \ac{PCT} is currently clamped.}
    \label{fig:robot}
    \vspace{-0.7cm}
\end{figure}
\section{Introduction} \label{sec:intro}
\acresetall
Enabling robots to act autonomously in uncertain environments has been a central topic of research in robotics for decades.
In many applications, not all relevant information about the object being manipulated is known, making the success of the manipulation skill highly uncertain. 
Even though considerable progress has been made, the success of robotic manipulation skills, such as grasping in uncertain environments as in industrial applications, remains a challenge. In order to address this uncertainty, researchers have focused on improving visual perception to achieve accurate results~\cite{marwan2021}. However, real-world scenarios in industries such as recycling present additional challenges, including changing light conditions, obstructed views, and vibrations that affect calibration, thus making visual feedback less reliable.

Automated disassembly of \ac{WEEE} is a complex and hazardous process, involving harsh working conditions and the manipulation of materials that pose a significant risk to human health. With global concerns mounting over environmental pollution and the efficient use of resources, the imperative for effective recycling and remanufacturing of \ac{WEEE} becomes ever more pressing. While robotic assembly processes are well-established and widely adopted in industry, research into automated disassembly has been limited~\cite{foumani.2020}. Notwithstanding the prediction that automated disassembly will become a crucial component of the industrial state-of-the-art within the next decade~\cite{poschmann.2020}, significant unanswered questions remain related to robotic disassembly, such as how to robustly detect unknown and damaged objects with altered shapes, rendering successful object manipulation difficult~\cite{vongbunyong.2017}. 

In this research paper, we focus on the problem of robotic grasping skills in uncertain environments, where the robot needs to act autonomously by perceiving its surroundings through haptics. The recycling of \ac{WEEE} products serves as a case study and an example for this problem.
In this context, geometric uncertainty remains the limiting factor and primarily arises from the inaccurate vision data used to compute the initially estimated grasping pose. We propose the use of inexpensive, conventional grippers in conjunction with an industrial robot (illustrated in Fig.~\ref{fig:robot}). In this case, its control relies exclusively on data obtained from a \ac{FT} sensor located on the wrist. This combination represents an intriguing example of fusing tactile and vision data to enhance industrial grasping strategies. The outcomes of this research have the potential to improve productivity, decrease the physical demands, and minimize the risks associated with repetitive work in \ac{WEEE} recycling. Our algorithms, however, are not limited to this application and can be applied to other areas with comparable industrial requirements, such as food handling and packaging by adjusting the hardware accordingly.

The key contributions of this research work are:
\begin{itemize}
\item Selection of suitable grippers based on cost-effectiveness, feasibility, robustness, and flexible grasping.
\item Integration of grasping tasks into a superordinate concept with task planner and vision unit, facilitating data exchange about objects and status updates.
\item Development of force-based strategies to improve robustness of grasping tasks in uncertain vision input scenarios, successfully demonstrated through disassembly of four different electronic devices (\ac{EL}, \ac{FPD}, \ac{MW}, and \ac{PCT}) using three different grippers in both lab and recycling plant environments.
\end{itemize}
The paper is organized as follows: First, we review related literature. Next, we present a detailed problem description, including the hardware selection that we employed to address the issue. We then provide an overview of the underlying methods and general background. Subsequently, we explain the tool-dependent grasping algorithms that we developed, and analyze and discuss their effectiveness in a laboratory setting and in an industrial recycling facility. Finally, we offer concluding remarks and discuss future research directions.
\subsubsection*{Notation:} Vectors $\boldsymbol{v} \in \mathbb{R}^n$ are denoted in lowercase bold font and matrices $M \in \mathbb{R}^{m \times n}$ in uppercase regular font. 
The base frame of our considered kinematic robotic chain is called $\{B\}$, the \ac{EE} frame is $\{EE\}$ and the \ac{TCP} frame is $\{TCP\}$. 
The coordinate transformation from base to \ac{EE} frame is $^BT_{EE}$ and consists of the rotation matrix $^B R_{EE}$ and the translational vector $^B \boldsymbol{p}_{EE}$. 
Robot wrenches (forces and torques) measured by the \ac{FT} sensor are denoted as $\boldsymbol{\mathcal{F}} \in \mathbb{R}^6$ and a twist in frame $\{a\}$ is presented as $\boldsymbol{\mathcal{V}}_a \in \mathbb{R}^6$.
\section{Related Work} \label{sec:relatedWork}
A general robotic recycling survey~\cite{bogue.2019} reveals that robots are mostly used for sorting tasks throughout the recycling process, as in for example \cite{guo.2021}.
Although some device-specific recycling robots have been developed~\cite{foo.2022}, their scope of application is limited, restricting industry-wide deployment.
Early attempts at designing robotic disassembly cells (such as~\cite{scholz.1999, Hohm.2000}) focused on specific devices or specialized tool development, but most of them failed to demonstrate successful experimental setups due to limitations in handling uncertainties commonly encountered during recycling and disassembly tasks~\cite{schmitt.2011}.
While some recent studies have attempted to address the issue of uncertainty in robotic grasping for disassembly processes by incorporating tactile and haptic data, these studies lack performance evaluation for real-world \ac{WEEE} devices~\cite{serrano.2021} or are limited in terms of flexible use for disassembly of multiple devices as they only consider one specific type~\cite{huang.2021}. 
The present study does not focus on visual identification, classification, sorting, or task planning methodologies, which are discussed in other studies such as \cite{kim.2006, wurster.2022} for product-driven control sequence generation and optimization of disassembly sequencing, respectively, and in \cite{mao2021, athanasiadis.2020} for visual identification and classification approaches.

Grasping in uncertain environments is a common challenge in many use-cases, including but not limited to \ac{WEEE} disassembly, and has been extensively studied. 
The uncertainty can be divided into two types \cite{wang2020}: physical uncertainty, which pertains to the weight and material properties of objects, and geometric uncertainty, which in the context of recycling poses significant challenges as the global position, pose, and dimensions of an object are not always obtainable. 
According to a survey by Marwan et al. \cite{marwan2021}, many researchers focus on improving visual detection to reduce geometric uncertainty. However, accurate visual detection is not always possible due to obstructed views or difficult calibration. To overcome this challenge, some approaches use tactile information to reduce uncertainty. In \cite{dragiev2013}, an exploration/exploitation approach is used to iteratively reduce uncertainty only relying on tactile data. However, these approaches often rely on assumptions about sensor availability or are limited to simulation results. Some approaches use both vision and haptic information to reduce uncertainty, but they are typically focused on specific objects, e.g., very thin objects \cite{levesque2018}, and require specialized grippers, e.g., a sensing soft pneumatic gripper \cite{chen2018}.

To effectively apply grasping strategies in recycling operations, there is a need to enhance flexibility in achieving grasping for various objects, and prioritize algorithms that can fuse initial vision and tactile feedback data in real-time. This must be achieved in a real-world industrial robot setup, with access to just one \ac{FT} sensor, which we identify as the key research gap in this domain.
\vspace{-0.2cm}
\section{Problem Description} \label{sec:problemDescription}
\begin{figure}[tb]
     \centering
      \begin{subfigure}[b]{0.15\textwidth}
         \centering
         \includegraphics[height=0.8\textwidth, trim={0 5cm 0  4cm},clip]{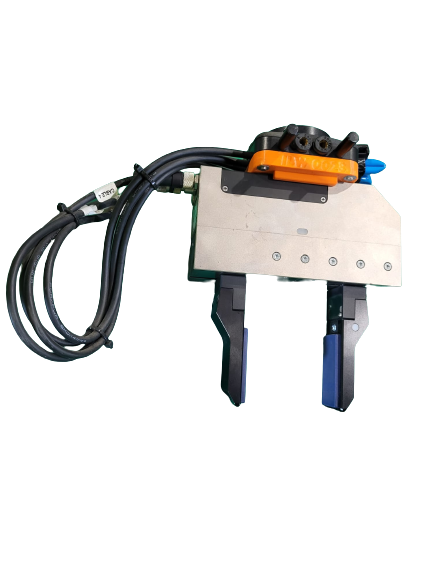}
         \caption{Tactile finger (A).}
         \label{fig:sensitivefinger}
     \end{subfigure}
    \hfill
     \begin{subfigure}[b]{0.15\textwidth}
         \centering
         \includegraphics[height=0.8\textwidth, trim={0 3cm 0  5cm},clip]{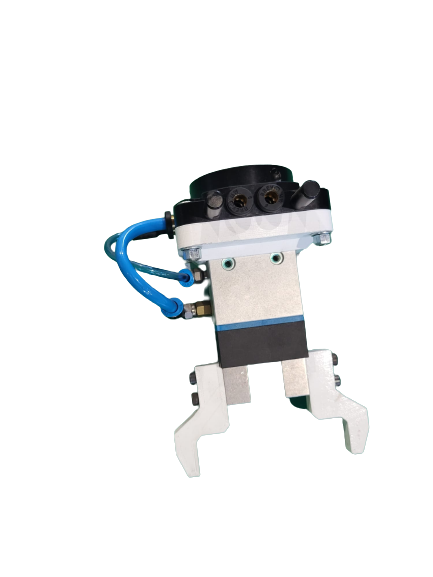}
         \caption{Slim finger (B).}
         \label{fig:thinfinger}
     \end{subfigure}
     \hfill
     \begin{subfigure}[b]{0.15\textwidth}
         \centering
         \includegraphics[height=0.8\textwidth, trim={0 5cm 0  5cm},clip]{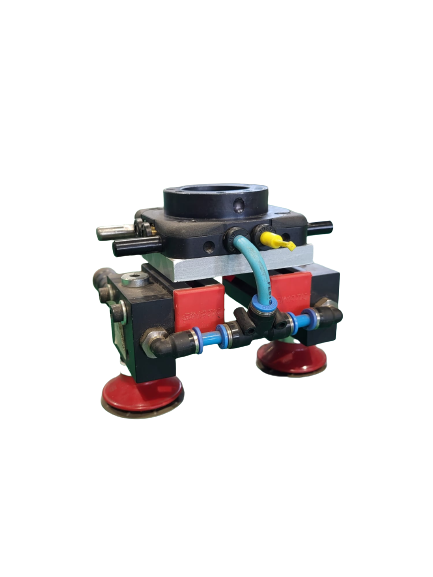}
         \caption{Vacuum (C).}
         \label{fig:surfacegripper}
     \end{subfigure}
        \caption{The selected grippers with tool changer side B attached.}
        \label{fig:grippers}
        \vspace{-0.6cm}
\end{figure}
The main challenges that remain to be addressed relate to the development of robust grasping strategies in the uncertain environment of a recycling cell. 
These challenges involve handling uncertainties in object pose, varying object sizes, shapes, and weights, and operating in a cluttered environment, all while minimizing the number of tools required to keep the recycling effective and easy to maintain. 
The robot must rely on a camera situated above the robot arm for detecting object features. 
As a result, during task execution, the robot does not receive any vision feedback and must depend solely on force feedback from the \ac{FT} sensor. 

The problem then involves finding a suitable pose $^B T_{\text{EE, des}}$ of the robot \ac{EE}, that enables successful grasping of an object. 
However, this desired grasping pose $^B T_{\text{EE, des}}$ is generally unknown. 
The objective of this study is to investigate how an industrial robot can adjust its \ac{EE} pose during the grasping skill, without relying on continuous visual feedback. 
Instead, the sensory input is limited to the external \ac{FT} data $\boldsymbol{\mathcal{F}}$.
The vision and task planner unit send the information about the grasping skill as,
\vspace{-0.3cm}
\begin{subequations}
\label{eq: xnd}
    \begin{align}
        \tilde{\boldsymbol{x}}_{\text{obj}} &= \boldsymbol{x}_{\text{obj}} + \boldsymbol{\sigma}_{x} \label{eq: x} \in \mathbb{R}^3 ,\\
        \tilde{\boldsymbol{n}}_{\text{obj}} &= \boldsymbol{n}_{\text{obj}} + \boldsymbol{\sigma}_{n} \label{eq: n} \in \mathbb{R}^3 ,\\
        \tilde{\boldsymbol{d}}_{\text{obj}} &= \boldsymbol{d}_{\text{obj}} + \boldsymbol{\sigma}_{d} \label{eq: d} \in \mathbb{R}^2 ,
    \end{align}
\end{subequations}
where $\boldsymbol{x}_{\text{obj}}$ is the unknown exact center of the object, $\boldsymbol{n}_{\text{obj}}$ is the unknown exact orientation of the object presented as a normal vector, $\boldsymbol{d}_{\text{obj}}$ is the exact unknown symmetric planar dimensions and $\boldsymbol{\sigma}_{x}$, $\boldsymbol{\sigma}_{n}$, $\boldsymbol{\sigma}_{d}$ denote the uncertainty for the location, orientation and dimensions of the object, respectively. Their distribution is unknown. $\tilde{\boldsymbol{x}}_{\text{obj}}$, $\tilde{\boldsymbol{n}}_{\text{obj}}$, and $\tilde{\boldsymbol{d}}_{\text{obj}}$, are then the noisy center, orientation and dimensions of the object, respectively. These uncertainties need to be reduced using tactile methods presented in this study.
\vspace{-0.2cm}
\section{Hardware} \label{sec:hardware}
In order to address the problem of adaptable and robust grasping in a recycling environment, we will examine the necessary hardware specifications and the choices made for this specific study.
\emph{Requirements:}
Designing cost-effective disassembly tools that are flexible and efficient is essential due to the significant challenge of cost effectiveness in disassembly, even in case of manual disassembly, as highlighted in~\cite{gkeleri.2008}.
Specialized tools for each task lead to increased costs and maintenance efforts, especially if not readily available off-the-shelf.
Additionally, tool-changing routines are time-consuming, reducing the overall efficiency of the disassembly process.
As such, the objective is to minimize the number of tools required while ensuring that they are robust enough to withstand forces exerted during manipulation of damaged devices and applicable to a wide range of objects and tasks. 
Another important consideration is the specifications for the entire disassembly cell, which must meet several requirements. The robot used in the cell should be suitable for industrial settings and equipped with collaborative features for safe interaction with humans.
The devices that are considered in this case study are mentioned in Section~\ref{sec:intro}. An overview of the components of these devices is outlined in Table~\ref{tab:objects}.
\begin{table*}[htb]
\centering
\caption{Objects to be grasped of the selected devices, their properties and the applicable gripper strategies. The grippers are specified in Fig. \ref{fig:grippers} and their strategies in Sec. \ref{sec:GripA} -- \ref{sec:gripC}.}
\begin{tabular}{|l|l|l|l|}
\hline
\textbf{Device}      & \textbf{Objects} & \textbf{Properties/Description}                                                                  & \textbf{Proposed Gripper (Strategy)} \\ \hline
\multirow{2}{*}{\ac{MW}}  & Cover            & large flat surface, metal                                                            & C                                  \\ \cline{2-4} 
                     & Magnetron        & small even surface, laminated structure on the side, metal                           & A or C                             \\ \hline
\multirow{3}{*}{\ac{PCT}} & Cover            & large flat surface, plastic                                                          & C                                  \\ \cline{2-4} 
                     & PSU              & box-shape of medium size, flat metal surfaces, very close to housing, heavy & C                                  \\ \cline{2-4} 
                     & PCU cooler       & medium size, small flat top surface, laminated structure on the side, metal          & A or C                             \\ \hline
\multirow{4}{*}{\ac{EL}}  & Cover (top)      & large, slightly bend but smooth surface, plastic                                     & C                                  \\ \cline{2-4} 
                     & Cover (middle)   & large and smooth surface with a groove in the middle                                 & C                                  \\ \cline{2-4} 
                     & Light bulb       & long cylinder, very delicate, rotation necessary to grasp, glass                     & A                                  \\ \cline{2-4} 
                     & Battery          & short cylinder, close to housing and in cluttered environment                        & B                                  \\ \hline
\multirow{2}{*}{\ac{FPD}} & Display Cover    & large flat surface, plastic                                                          & C                                  \\ \cline{2-4} 
                     & Display          & large flat surface, heavy                                                            & C                                  \\ \hline
\end{tabular}
\label{tab:objects}
\vspace{-0.5cm}
\end{table*}
\emph{Selection:}
The robot chosen is an industrial COMAU Racer~5 robot. 
For the flexible task of recycling it needed to be equipped with a \ac{FT} sensor and a tool-changer as shown in Fig.~\ref{fig:robot}. 
Two types of grippers were considered: finger-based and vacuum-based suction grippers. 
We propose using tools that are easy to handle, control, repair, and replace and can be designed to disassemble many different objects in \ac{WEEE}. 
Specifically, we selected a gripper with sensitive fingers for handling delicate objects (Fig.~\ref{fig:sensitivefinger}), a gripper with slim fingers to reach objects in cluttered environments (Fig.~\ref{fig:thinfinger}), and a surface gripper for large objects with flat surfaces (Fig.~\ref{fig:surfacegripper}).
\section{Methods \& Setup Description} \label{sec:preliminaries}
In industrial settings, manipulation skills cannot be considered as isolated abilities. 
Hence, we will provide a brief overview of how these skills fit into the broader recycling process and how the inputs for the system are generated. 
We will also describe how we calculate the desired poses using visual inputs and provide a brief explanation of the control architecture employed on the robot to grasp objects.
\subsection{Embedding of Grasping Skills in the Recycling Line}\label{sec:embedGraspSkills}
\begin{figure}[tb]
    \centering
    \includegraphics[width =\linewidth, trim={0 13cm 0 0},clip]{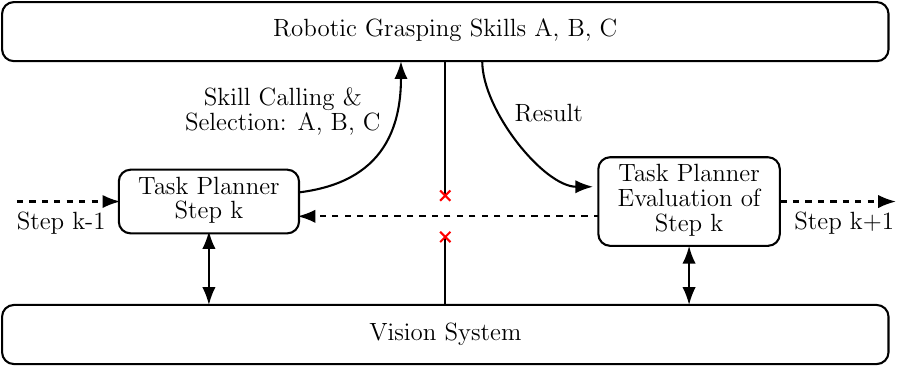}
    \caption{Embedding of the robotic grasping skills in the overall disassembly procedure. Our focus is on the top box containing the robotic grasping skills, which can be called by the task planner and delivers a result back to it. There is no feedback loop between the vision and robotic units in this system. 
}
    \label{fig:general}
    \vspace{-0.6cm}
\end{figure}
The problem setting is depicted in Fig.~\ref{fig:general}, which illustrates the embedding of the robotic grasping skills in the overall recycling procedure. 
Although this paper does not cover the task planner and vision unit in detail, we will provide a brief overview of their functionality since we assume communication between them and our study.
The task planner operates as a state machine. Assuming a state $\mathrm{k}$ (as shown in Fig.~\ref{fig:general}) where the task is to remove an object by grasping it, the task planner requests the vision system to identify and classify the object, as well as provide its estimated center and orientation.
Based on this information, the task planner selects a grasping strategy and transmits the vision information to the robotic system.
At this point, the robotic system must rely on its own haptic exploration strategies, based on the selected gripper, to successfully pick up the object. 
Once the skill is completed, the robotic platform returns the result to the task planner, which then determines the next action to be taken.
If the removal is successful, the task planner asks the vision system for confirmation. 
If there is a conflict or a failed skill result, the task planner evaluates the other received information to determine whether the step should be repeated or if a human operator needs to be called for assistance.
\vspace{-0.2cm}
\subsection{Computing the Desired \ac{EE} Pose}\label{sec:computePose}
In the grasping skills, we need to determine the desired grasping pose for the tool 
\begin{equation}
^\text{B}T_{\text{TCP, des}} = 
\begin{bmatrix}
&^\text{B}R_{\text{TCP, des}} & & \boldsymbol{p}_{\text{trans}} \\
0&0 &0                   & 1 \\
\end{bmatrix} \in \mathbb{R}^{4 \times 4} ,
\label{eq:goalpose}
\end{equation}
where $^\text{B}R_{\text{TCP, des}} \in \mathbb{R}^{3 \times 3}$ is the rotation matrix and $\boldsymbol{p}_{\text{trans}}\in \mathbb{R}^3$ is the translation vector of the pose for the tool tip. From the task planner input we only obtain 3 vectors~\eqref{eq: xnd}.
As we want the tool to grasp the object, we set $\boldsymbol{p}_{\text{trans}} = \boldsymbol{x}_{\text{obj}}$.
Traditionally, as stated in~\cite{corke2011}, the axis of the tool corresponds with the z-axis and is referred to as the approach vector, represented by $\hat{\boldsymbol{a}} = [\hat{a}_x, \hat{a}_y, \hat{a}_z]^\mathsf{T}$, where, in our case $\hat{\boldsymbol{a}} = -\boldsymbol{n}_{\text{obj}}$.
However, specifying solely the z-axis does not adequately describe the coordinate frame; it is also necessary to indicate the direction of the x-axis and y-axis. 

An orientation vector that is perpendicular and provides orientation, such as between the two fingers of a gripper, is introduced as $\hat{\boldsymbol{o}} = [\hat{o}_x, \hat{o}_y, \hat{o}_z]^\mathsf{T}$. 
In our case, all devices are pre-sorted and placed on the table by a second robot in the disassembly which is why all objects to be grasped are aligned with the x-axis or y-axis up to an alignment error arising from damage to the devices or imprecise clamping.
We  obtain $\hat{\boldsymbol{o}}$ by evaluating the dimensions of the object $\boldsymbol{d}_{\text{obj}}$. 
In other scenarios, where the orientation cannot be guaranteed to be along an axis, the same approach is usable but the orientation vector must also be obtained by an external observer. 
These two unit vectors are sufficient to define the rotation matrix,
\begin{equation}
    ^\text{B}R_{\text{TCP, des}} = \begin{bmatrix}
    n_x &o_x &a_x \\
    n_y &o_y &a_y \\
    n_z &o_z &a_z
    \end{bmatrix},
\end{equation}
for $\boldsymbol{n} = \hat{\boldsymbol{o}} \times \hat{\boldsymbol{a}}$, $\boldsymbol{a} = \hat{\boldsymbol{a}}$ and $\boldsymbol{o} = \boldsymbol{a} \times \boldsymbol{n}$. 
The desired \ac{EE} goal pose is given by $^\text{B}T_{\text{EE, des}} = \  ^\text{B}T_{\text{TCP}} \ ^\text{TCP}T_{\text{EE, des}}$. 
\vspace{-0.1cm}
\subsection{Hybrid Control Strategy} \label{controlStrategy}
While the COMAU robot is not typically controlled via motor torque or impedance-based control interfaces, it does allow for the position of the \ac{EE} $\boldsymbol{x}$ to be controlled in the form of a Cartesian deviation relative to the current \ac{EE} position. 
To enable compliant manipulation tasks, the control interfaces of this industrial robot are customized to use a Cartesian hybrid force-velocity controller. In this work, we use a controller based on \cite{khatib.1986, craig.1979} which is explained in detail in \cite{gabler.2022a, gabler.2022b}. 
The controlled system can be simplified to 
$\boldsymbol{x}_{t+1} = \boldsymbol{x}_t + \boldsymbol{\delta}_x = \boldsymbol{x}_t + \displaystyle{\boldsymbol{u}_{\boldsymbol{\mathcal{V}}_{\text{EE, des}}}}\delta_t$, 
where $\boldsymbol{\delta}_x$ represents the control command sent to the robot. Since the robot runs at a safe, constant update rate $\delta_t$ in real time, the Cartesian deviation command $\displaystyle{\boldsymbol{u}_{\boldsymbol{\mathcal{V}}_{\text{EE, des}}}}$ can also be used to send a Cartesian forward velocity command to the robot. To achieve a hybrid force-velocity control policy for the robot system, this forward \ac{EE} velocity follows a hybrid Cartesian force-velocity controller \cite{khatib.1986}
\begin{subequations}
\label{eq: controller}
    \begin{align}
        \boldsymbol{u}_{\boldsymbol{\mathcal{V}}_{\text{EE, des}}} &= S_{\text{vel}} \boldsymbol{\mathcal{V}}_{\text{EE, des}} + S_{\text{frc}} K_P (\boldsymbol{\mathcal{F}}_{des} - \boldsymbol{\mathcal{F}}) \label{eq:control1}\\
        &= S_{\text{vel}} \boldsymbol{s} \boldsymbol{\mathcal{V}}_{\text{EE, max}} + S_{\text{frc}} K_P (\boldsymbol{\mathcal{F}}_{des} - \boldsymbol{\mathcal{F}}),
        \label{eq:control2}
    \end{align}
\end{subequations}
where $\boldsymbol{s}$ is a scaling vector given the maximum \ac{EE} twist $\boldsymbol{\mathcal{V}}$ and $K_P$ is a positive definite proportional control gain matrix. 
The selection matrices $S_{\text{frc}}$ and $S_{\text{vel}}$, used to activate velocity and force control modalities along selective axes in \eqref{eq: controller}, are specified in \cite{gabler.2022b}. All control goals are sent in the \ac{EE} coordinate frame which means the desired wrenches and twists in the TCPs for all tools with different $^\text{EE}T_{\text{TCP}}$ need to be determined analytically.
The transformations \cite{lynch2017} are given as
\begin{subequations}
\label{eq: twistandwrench}
    \begin{align}
        \boldsymbol{\mathcal{V}}_{\text{EE, des}} &= \text{Ad}_{^\text{EE}T_{\text{TCP}}}^\mathsf{T}  \boldsymbol{\mathcal{V}}_{\text{TCP, des}} ,
    \label{eq:Vtransform}\\
        \boldsymbol{\mathcal{F}}_{\text{EE, des}} &= \text{Ad}_{^\text{EE}T_{\text{TCP}}}^\mathsf{T}  \boldsymbol{\mathcal{F}}_{\text{TCP, des}},
        \label{eq:Wtransform}
    \end{align}
\end{subequations}
where $\boldsymbol{\mathcal{V}}_{\text{EE, des}}$, $\boldsymbol{\mathcal{V}}_{\text{TCP, des}}$, $\boldsymbol{\mathcal{F}}_{\text{EE, des}}$ and $\boldsymbol{\mathcal{F}}_{\text{TCP, des}}$ are the desired twists and wrenches in the \ac{EE} frame and the TCP frame, respectively and the adjoint presentation of $^\text{EE}T_{\text{TCP}}$ is $\displaystyle{\text{Ad}_{^\text{EE}T_{\text{TCP}}}}$.
To avoid any potential collisions, all movements of the robot are planned using a motion planner~\cite{bari2021}.
\subsection{Safety in a Shared Workspace} \label{sec:workSpaceSafety}
Since the current setup is a collaborative disassembly setup, where some tasks are taken over by a human worker, it is necessary to call in a human operator. 
This may be the case when defects are detected or for parts that cannot be disassembled by the robot. 
For the experiment setup on the industrial factory floor (see Section~\ref{sec:analysisDiscussion}), the cell is equipped with safety sensors that are connected to the internal driver of the robot and also to a camera that publishes the current safety status via the Robotic Operating System (ROS). 
In the software implementation, a safety check is run in parallel to all commands at all times, which includes the evaluation of the camera and sensor information.
A triggered stop results in an immediate halt of robot and tool movements, with resumption only after safety area clearance. When entering an alert area, the robot reduces its maximum speed and restores it only after clearance.
\vspace{-0.1cm}
\section{Grasping Strategies under Uncertainty} \label{sec:graspingStrategies}
\vspace{-0.05cm}
\begin{figure}[htb]
    \centering
    \includegraphics[width = 0.45 \textwidth, trim={0 0.3cm 0 0},clip]{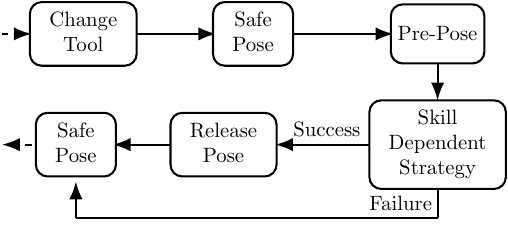}
    \caption{Common structure of all grasping skills. First, the system performs an automated tool change routine if needed. Afterwards, the robot moves to a safe position. The grasping algorithm determines a safe pre-pose and sends the robot to this pose.
    Then, the tool dependent grasping strategy is performed. If the grasping skill fails, the robot sends a \emph{failed} message to the task planner and returns to the safe pose to await the next call. 
    Otherwise, the robot delivers the object to a specified release pose, before returning to a safe position.}
    \label{fig:allskills}
    \vspace{-0.2cm}
\end{figure}
A request sent by the task planner to the robotic platform includes information on the most suitable grasping strategy, A, B, or C, along with an estimate of the center position, orientation, and dimensions of the object \eqref{eq: xnd}.
The called grasping skills all share a common structure, which is summarized in Fig.~\ref{fig:allskills}. The mentioned secure pre-pose $^\text{B}T_{\text{TCP, pre}} = d_{\text{safety}} \ ^\text{B}\tilde{\boldsymbol{n}}_{\text{obj}} \ ^\text{B}T_{\text{TCP, des}}$ is adjacent to the estimated object position $\tilde{\boldsymbol{x}}_{\text{obj}}$, along the estimated normal vector $\tilde{\boldsymbol{n}}_{\text{obj}}$, at a safe distance $d_{\text{safety}}$ and is calculated by the grasping algorithm. The safe distance $d_{\text{safety}}$ is determined by the object and gripper characteristics, as well as estimated levels of uncertainty $\boldsymbol{\sigma}_{x}$, $\boldsymbol{\sigma}_{n}$, $\boldsymbol{\sigma}_{d}$ in the vision data~\eqref{eq: xnd}. In the following, the proposed grasping strategies will be described.
\vspace{-0.1cm}

\subsection{Grasping Strategy A: Tactile Finger Gripper}
\label{sec:GripA}
\begin{figure}
    \centering
    \includegraphics[scale=0.28]{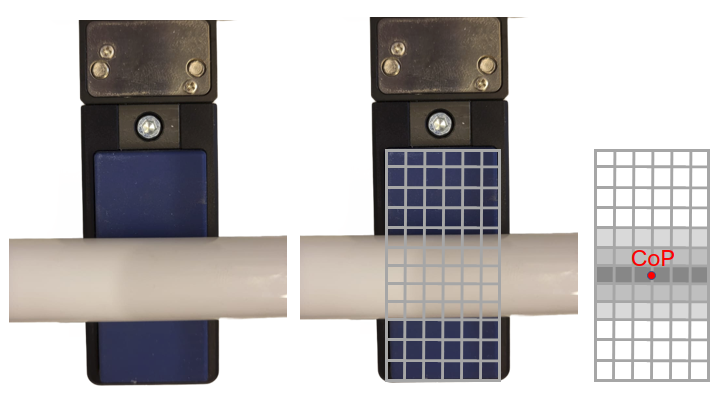}
    \caption{Tactile finger gripper with the \ac{EL} light bulb pressing against it, the resulting pressure distribution on the sensor array and the calculated CoP.}
    \label{fig:adgrip}
    \vspace{-0.5cm}
\end{figure}

In order to handle delicate objects such as a light bulb in an \ac{EL}, a grasping approach that is highly sensitive to incorrect poses is required. 
Therefore, we propose the use of a gripper with sensitive fingers that uses a pressure sensor array as shown in Fig.~\ref{fig:adgrip}. 
This approach is an adaptation of~\cite{gabler.2022a} for the use in robotic disassembly tasks.
The control method uses the force feedback of the sensor array of the fingers additionally to the one seen on the \ac{FT} sensor and guarantees compliance in defined directions based on the pressure distribution seen on the fingers. 
With the sensitive fingers, we can accurately determine the \ac{CoP}, allowing to rotate the object around a certain axis of the obtained point. 
This feature is crucial for example for disassembling the \ac{EL}, where the light bulb requires rotation by a few degrees.

Similar to~\eqref{eq:Vtransform}, the desired twist of the \ac{EE}, denoted as $\mathcal{V}_{\text{EE, des}}$, is now dependent on the desired twist of the \ac{CoP}. 
The point of rotation is no longer fixed at the TCP, and instead, it needs to be dynamically determined based on the pressure distribution. 
Thus, $\text{Ad}_{T_{\text{EE,CoP}}}^\mathsf{T}$ becomes a variable that changes over time with the changes in $T_{\text{EE,CoP}}$. 
By using the method proposed in~\cite{gabler.2022a} to calculate the current \ac{CoP}, we use that the changes of $^\text{EE}T_{\text{CoP}}$ lay in its translational part $^\text{EE}\boldsymbol{p}_{\text{CoP}} = [0, 0, \ ^\text{EE}z_{\text{CoP}}]^\mathsf{T}$ ,
where $^\text{EE}z_{\text{CoP}}$ is the total distance from the EE to the \ac{CoP} along the z-axis of the EE coordinate system, meaning that it is given as $^\text{EE}z_{\text{CoP}} = \ ^\text{EE}z_{\text{TCP}} - \ ^\text{TCP}z_{\text{CoP}}$ if the TCP is at the tip of the gripper.

\begin{figure}[tb]
    \centering
    \includegraphics[width =0.93\linewidth]{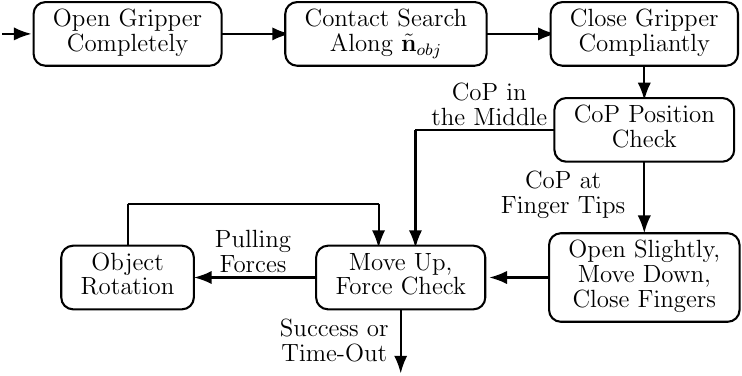}
    \caption{Tactile finger grasping procedure.}
    \label{fig:adaptive_diagram}
    \vspace{-0.7cm}
\end{figure}

We propose the grasping procedure summarized in Fig.~\ref{fig:adaptive_diagram} for this gripper. 
Initially, the gripper starts fully open and moves towards the goal pose until it makes contact with the object or reaches a small distance beyond the goal pose.  Then, the gripper is closed while being compliant around its z-axis and in its x-direction and y-direction until the force difference on both sides and the force on the \ac{FT} sensor are small. 
If the \ac{CoP} is found to be only at the tip of the fingers, the fingers are slightly opened again and the gripper is moved along the normal vector for another small distance (1 cm for this type of gripper) or until contact is made.
Subsequently, the gripper is closed and the robot attempts to move back to the pre-pose while holding the object. 
If high pulling forces are detected, the motion is terminated, and it is assumed that the object is still in a fixture.
In such cases, an attempt to rotate the object around the CoP is started.
If successful, it is concluded there is a rotatory joint; otherwise, a different kind of fixture is encountered, and a \emph{failed} message is sent. 
In the case of the light bulb, the object will be lifted again after rotation. 
If the lifting fails again, the fingers are opened, moved back to the upright position, closed again, and the rotating process is repeated. 
This sequence continues until a \emph{time-out} or successful object lifting occurs. 
The process can be expedited if preliminary information about the object is available. 
For instance, the initial opening width can be reduced, and we can anticipate whether a rotatory joint is present.
\subsection{Grasping Strategy B: Slim-Fingered Gripper}
\label{sec:gripB}
\begin{figure}[htb]
    \centering
    \includegraphics[width=\linewidth]{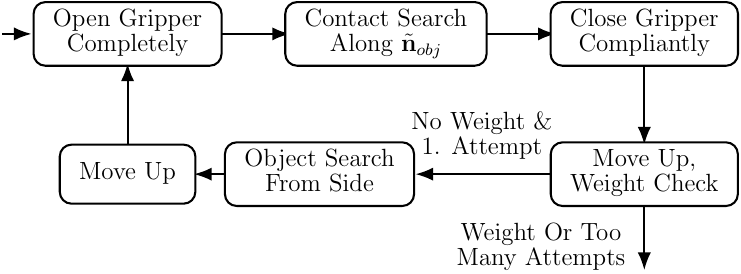}
    \caption{Grasping procedure of slim-fingered gripper.}
    \label{fig:nonsensitive}
    \vspace{-0.5cm}
\end{figure}

Although the tactile finger gripper can perform various grasping tasks with different objects, its size make it unsuitable for certain tasks and poses. In order to grasp inside narrow spaces, a smaller gripper is required. In the present work, a pneumatic gripper with 3d printed fingers was used, see Figure~\ref{fig:thinfinger}.
An outline of the proposed procedure for this gripper is presented in Figure~\ref{fig:nonsensitive}. The initial step involves fully opening the gripper as the robot approaches the desired pre-pose. The robot then follows the provided normal vector to move towards the object and compliantly closes the fingers upon contact. The first attempt to remove the object is made, and a weight check is carried out. If the weight check detects an object, the robot delivers it to the release pose. However, if no object is detected, the grasping attempt is deemed unsuccessful, requiring a more precise goal position. To achieve this, a search approach is suggested, mimicking human behavior. The gripper moves next to the estimated object position, for example by a distance of $1.2 \ l_{width}$, where $l_{width}$ is the opening width. Then, the gripper slowly approaches the estimated object position until contact is made. Contact with the object reveals its exact location. If this approach fails, the same process can be attempted on the other side. This process allows for accurate positioning of the gripper to enable successful grasping of the object despite the limited opening width. This procedure can be applied to the tactile gripper (A), although it was not required for the objects and devices considered in this study since the opening width was significantly wider than the object width.
\subsection{Grasping Strategy C: Vacuum Gripper}
\label{sec:gripC}
The two finger grippers are suitable for grasping small objects with various shapes, but they are limited in handling larger objects due to their opening width. To overcome this limitation, a vacuum gripper is introduced, which is particularly useful for flat objects that are too large for the finger grippers or too close to obstacles. This gripper can reach objects from above, regardless of their exact dimensions, as long as there is enough space for at least one suction cup.

\begin{figure}[b]
\vspace{-0.5cm}
    \centering
    \includegraphics[width=\linewidth]{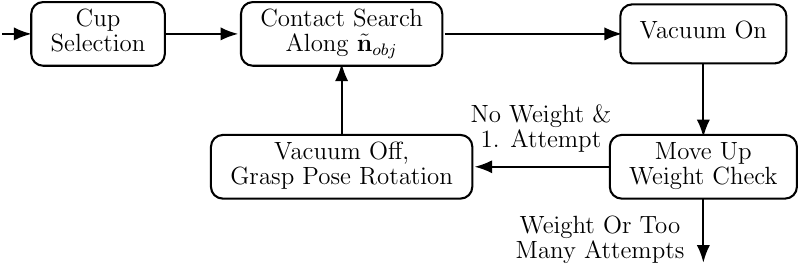}
    \caption{Grasping procedure for vacuum gripper.}
    \label{fig:vacuum}
    \vspace{-0.5cm}
\end{figure}

The procedure for using the gripper is summarized in Fig.~\ref{fig:vacuum}. First, the suction cup to be used is decided based on the object dimensions $\tilde{\boldsymbol{d}}$. If both cups can be used which is beneficial for stability, the gripper centers itself on the object; otherwise, one side is selected, and the target position is adjusted accordingly. The gripper then moves to the pre-pose, descends along the normal vector until contact is made, turns on the vacuum, and moves up to check for a change in weight. The gripper orientation is crucial for task success as it can collide with obstacles. Therefore, if the task fails, we propose rotating the gripper 45 $^{\circ}$, either three times around the center if both cups are used, or seven times around the selected cup. The angles are heuristically chosen to balance time and accuracy and can be adapted according to the needs.
\vspace{-0.2cm}
\subsection{Comparison of the Grasping Strategies}
\begin{table*}[htb]
\centering
\caption{Summary and comparison of grasping skills for different object properties, divided into the cases when accurate vision data is provided (AV) and when only inaccurate, noisy data is available (IV). While the first three rows refer to standard grippers and non-intelligent skills, the last three highlighted rows represent our proposed grippers and skills.}
\label{tab:Summary}
\begin{tabular}{l|cc|cc|cc|cc|cc|cc|ll}

\cline{2-15}

& \multicolumn{2}{c|}{\rotatebox[origin=c]{90}{\textbf{\begin{tabular}[c]{@{}c@{}}Small \\ Objects\end{tabular}}}}
& \multicolumn{2}{c|}{\rotatebox[origin=c]{90}{\textbf{\begin{tabular}[c]{@{}c@{}}Medium\\ Sized \\ Objects\end{tabular}}}}
& \multicolumn{2}{c|}{\rotatebox[origin=c]{90}{\textbf{\begin{tabular}[c]{@{}c@{}}Large \\ Objects\end{tabular}}}}
& \multicolumn{2}{c|}{\rotatebox[origin=c]{90}{\textbf{\begin{tabular}[c]{@{}c@{}}Flat Surface \\ Inside an Object\\ Inaccessible Sides \\ (Small)\end{tabular}}}}   
& \multicolumn{2}{c|}{\rotatebox[origin=c]{90}{\textbf{\begin{tabular}[c]{@{}c@{}}Flat Surface \\ Inside an Object\\ Inaccessible Sides \\ (Large) \end{tabular}}}}  
& \multicolumn{2}{c|}{\rotatebox[origin=c]{90}{\textbf{\begin{tabular}[c]{@{}c@{}}Any Shape with \\ 1 Accessible \\ Flat Surface\end{tabular}}}}   
& \multicolumn{2}{c|}{\rotatebox[origin=c]{90}{\textbf{Any Shape}}}
                                                               \\

\cline{2-15} 
& \multicolumn{1}{c|}{\textbf{AV}} & \textbf{IV} & \multicolumn{1}{c|}{\textbf{AV}}& \textbf{IV}& \multicolumn{1}{c|}{\textbf{AV}} & \textbf{IV} & \multicolumn{1}{c|}{\textbf{AV}}     
& \textbf{IV} & \multicolumn{1}{c|}{\textbf{AV}}& \textbf{IV}  & \multicolumn{1}{c|}{\textbf{AV}} & \textbf{IV} & \multicolumn{1}{l|}{\textbf{AV}} & \multicolumn{1}{l|}{\textbf{IV}}\\ 

\hline

\multicolumn{1}{|l|}{\textbf{Standard 2-Finger}}                                                            
& \multicolumn{1}{c|}{\greencheck}                         
& \redcross                         
& \multicolumn{1}{c|}{\greencheck}                              
& \redcross                                                                          
& \multicolumn{1}{c|}{(\orangecheck)$^1$ }                          
& \redcross                                                                          
& \multicolumn{1}{c|}{\redcross}                                        
& \redcross                                       
& \multicolumn{1}{c|}{\redcross}                                       
& \redcross                                      
& \multicolumn{1}{c|}{\redcross}                                                                           
& \redcross                                 
& \multicolumn{1}{l|}{\redcross}                         
& \multicolumn{1}{l|}{\redcross}                         \\ 

\hline 

\multicolumn{1}{|l|}{\textbf{\begin{tabular}[l]{@{}l@{}}Standard Suction\\ (1 Cup)\end{tabular}}}         
& \multicolumn{1}{c|}{\redcross}   
& \redcross 
& \multicolumn{1}{c|}{(\orangecheck)$^{2,4}$ } 
& \multicolumn{1}{c|}{(\orangecheck)$^{2,4}$ } 
& \multicolumn{1}{c|}{\redcross}                          
& \redcross                                                                          
& \multicolumn{1}{c|}{\greencheck}              
& \greencheck             
& \multicolumn{1}{c|}{\redcross}                                       
& \redcross                                      
& \multicolumn{1}{c|}{(\orangecheck)$^4$ }                                       
& (\orangecheck)$^{3,4}$  
& \multicolumn{1}{l|}{\redcross} 
& \multicolumn{1}{l|}{\redcross} \\ 

\hline

\multicolumn{1}{|l|}{\textbf{\begin{tabular}[l]{@{}l@{}}Standard Suction\\ (Multiple Cups)\end{tabular}}} 
& \multicolumn{1}{c|}{\redcross}   
& \redcross & \multicolumn{1}{c|}{(\orangecheck)$^2$ } 
& \multicolumn{1}{c|}{(\orangecheck)$^2$ }  
& \multicolumn{1}{c|}{(\orangecheck)$^2$ } 
& \multicolumn{1}{c|}{(\orangecheck)$^2$ }  
& \multicolumn{1}{c|}{\redcross}                                        
& \redcross                                       
& \multicolumn{1}{c|}{\greencheck}             
& \redcross                                      
& \multicolumn{1}{c|}{(\orangecheck)$^5$ } 
& (\orangecheck)$^{3,5}$  
& \multicolumn{1}{l|}{\redcross} & \multicolumn{1}{l|}{\redcross} \\ 

\hline

\rowcolor[HTML]{C0C0C0} 
\multicolumn{1}{|l|}{\cellcolor[HTML]{C0C0C0}\textbf{A: Tactile Finger}}& \multicolumn{1}{c|}{\cellcolor[HTML]{C0C0C0}\greencheck} & \greencheck& \multicolumn{1}{c|}{\cellcolor[HTML]{C0C0C0}\greencheck}      & \greencheck  & \multicolumn{1}{c|}{\cellcolor[HTML]{C0C0C0}(\orangecheck)$^1$ }  & (\orangecheck)$^1$                                                                     & \multicolumn{1}{c|}{\cellcolor[HTML]{C0C0C0}\redcross}                & \redcross                                       & \multicolumn{1}{c|}{\cellcolor[HTML]{C0C0C0}\redcross}               & \redcross                                      & \multicolumn{1}{c|}{\cellcolor[HTML]{C0C0C0}\redcross}                                                   & \redcross                                 & \multicolumn{1}{l|}{\cellcolor[HTML]{C0C0C0}\redcross} & \multicolumn{1}{l|}{\cellcolor[HTML]{C0C0C0}\redcross} \\ 

\hline

\rowcolor[HTML]{C0C0C0} 
\multicolumn{1}{|l|}{\cellcolor[HTML]{C0C0C0}\textbf{B: Slim Finger}}
& \multicolumn{1}{c|}{\cellcolor[HTML]{C0C0C0}\greencheck} 
& \greencheck & \multicolumn{1}{c|}{\cellcolor[HTML]{C0C0C0}(\orangecheck)$^1$ }  
& (\orangecheck)$^1$   & \multicolumn{1}{c|}{\cellcolor[HTML]{C0C0C0}\redcross}        
& \redcross                                                                      
& \multicolumn{1}{c|}{\cellcolor[HTML]{C0C0C0}\redcross}                
& \redcross                                       
& \multicolumn{1}{c|}{\cellcolor[HTML]{C0C0C0}\redcross}               
& \redcross                                      
& \multicolumn{1}{c|}{\cellcolor[HTML]{C0C0C0}\redcross}                                                   
& \redcross                                 
& \multicolumn{1}{l|}{\cellcolor[HTML]{C0C0C0}\redcross} & \multicolumn{1}{l|}{\cellcolor[HTML]{C0C0C0}\redcross} \\
\hline

\rowcolor[HTML]{C0C0C0} 
\multicolumn{1}{|l|}{\cellcolor[HTML]{C0C0C0}\textbf{C: Suction}}                                        
& \multicolumn{1}{c|}{\cellcolor[HTML]{C0C0C0}\redcross}   
& \redcross                         
& \multicolumn{1}{c|}{\cellcolor[HTML]{C0C0C0}(\orangecheck)$^2$ } 
& \multicolumn{1}{c|}{\cellcolor[HTML]{C0C0C0}(\orangecheck)$^2$ }                 
& \multicolumn{1}{c|}{\cellcolor[HTML]{C0C0C0}(\orangecheck)$^2$ } 
& \multicolumn{1}{c|}{\cellcolor[HTML]{C0C0C0}(\orangecheck)$^2$ } 
& \multicolumn{1}{c|}{\cellcolor[HTML]{C0C0C0}\greencheck}              
& \cellcolor[HTML]{C0C0C0}\greencheck             
& \multicolumn{1}{c|}{\cellcolor[HTML]{C0C0C0}\greencheck}             
& \cellcolor[HTML]{C0C0C0}\greencheck            
& \multicolumn{1}{c|}{\cellcolor[HTML]{C0C0C0}\greencheck}                                                 
& \cellcolor[HTML]{C0C0C0}(\orangecheck)$^3$       
& \multicolumn{1}{l|}{\cellcolor[HTML]{C0C0C0}\redcross} 
& \multicolumn{1}{l|}{\cellcolor[HTML]{C0C0C0}\redcross} \\

\hline

\multicolumn{13}{l}{\begin{tabular}[c]{@{}l@{}} $^1$ Depending on the chosen dimensions (opening width) of the gripper and the objects. $^2$ If surface is flat. \\ $^3$ If flat surface is big enough to allow for some inaccuracy. $^4$ If object is not heavy. $^5$ If surface is big enough to accommodate all cups.\end{tabular}}   

\end{tabular}
\vspace{-0.3cm}
\end{table*}
Our study proposes three grasping algorithms that improve the usability of grippers in uncertain environments by incorporating tactile feedback to reduce uncertainty. These skills outperform standard grippers that rely solely on visual data, enabling more robust and efficient object manipulation while reducing the need for repeated grasp attempts with new vision data. Table \ref{tab:Summary} presents a comparison between standard and intelligent grippers, highlighting the benefits of our approach. Although there is still scope for enhancing grasping of objects with complex shapes, our approach of combining simple and affordable grippers with intelligent search techniques results in improved overall applicability. Overall, our study suggests that incorporating these grasping skills can significantly enhance the capabilities of robotic systems operating in uncertain environments in and beyond \ac{WEEE} recycling,


\vspace{-0.1cm}
\section{Validation and Discussion} 
\label{sec:analysisDiscussion}
\begin{figure*}[htb]
     \centering
    \hspace{2cm}
     \begin{subfigure}[b]{0.15\textwidth}
         \centering
         \includegraphics[height=1.6\textwidth, trim={0 1cm 10cm 0},clip]{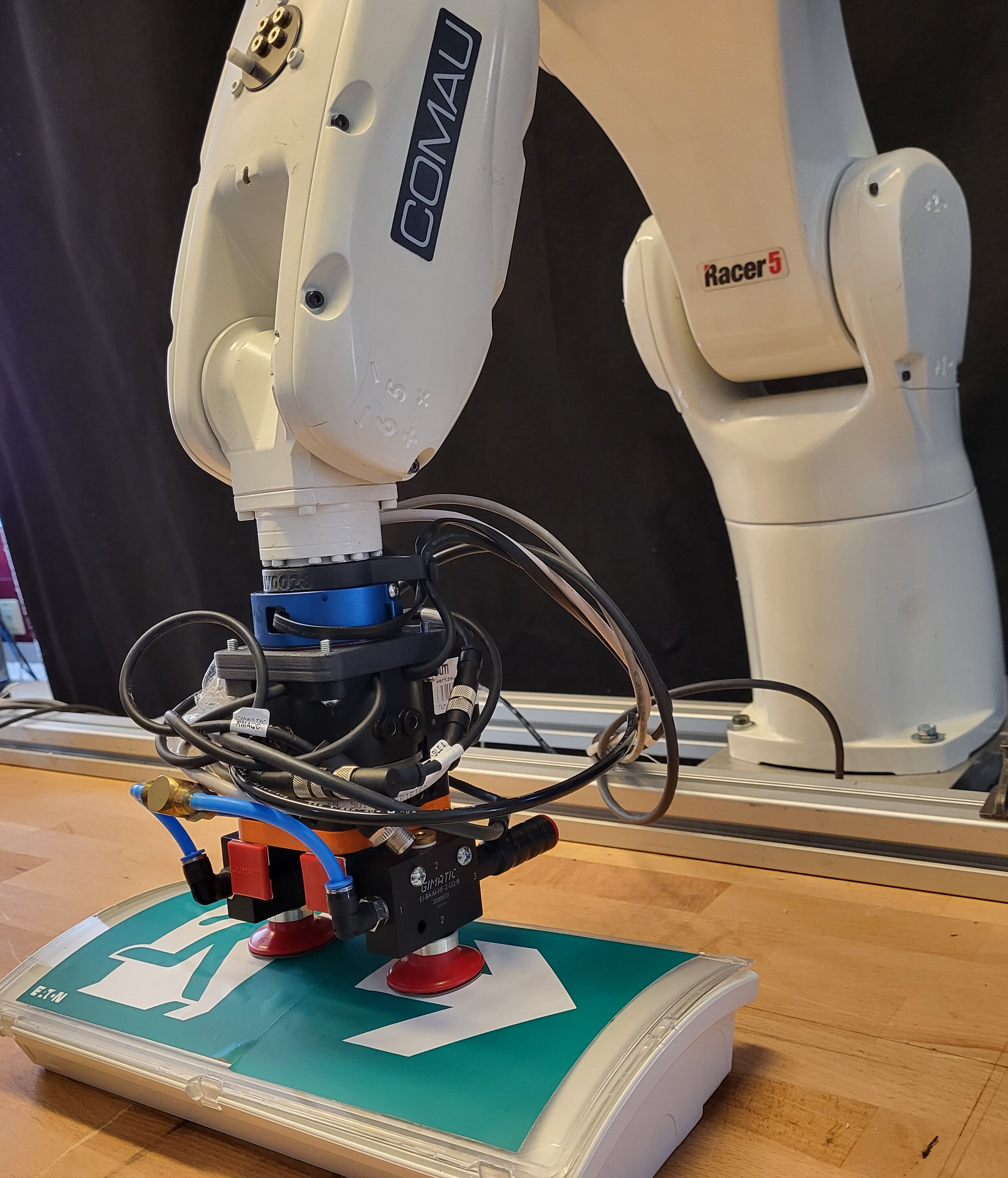}
         \caption{Cover I (C).}
         \label{fig:topcov}
     \end{subfigure}~
     \hspace{0.4cm}
     \begin{subfigure}[b]{0.15\textwidth}
         \centering
         \includegraphics[height=1.6\textwidth, trim={0 0 0 0},clip]{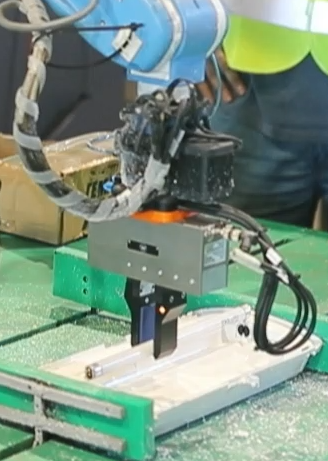}
         \caption{Lightbulb (A).}
         \label{fig:light}
     \end{subfigure}~
     \hspace{0.15cm}
     \begin{subfigure}[b]{0.15\textwidth}
         \centering
         \includegraphics[height=1.6\textwidth, , trim={0 0 3cm  30cm},clip]{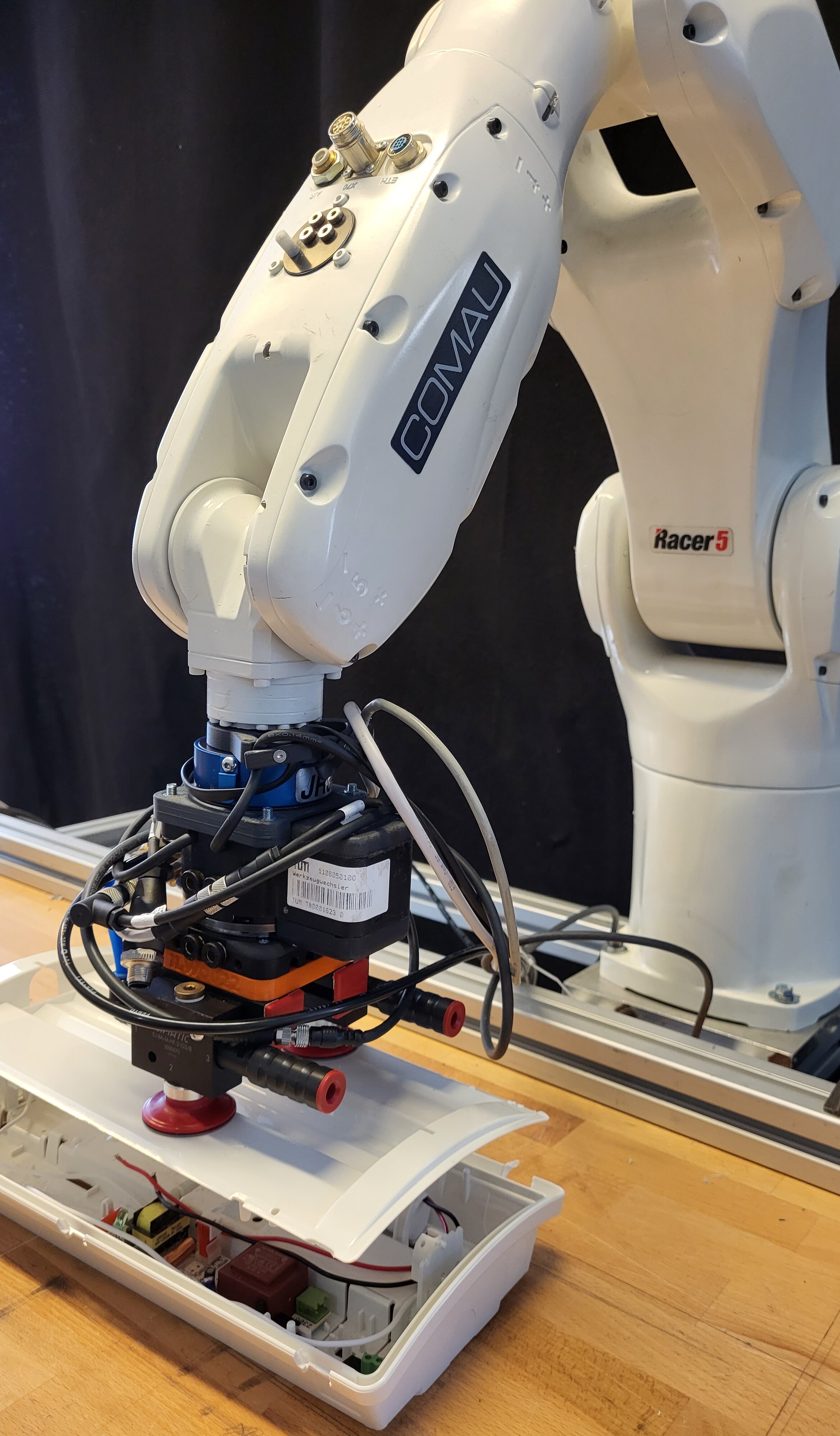}
         \caption{Cover II (C).}
         \label{fig:midcov}
     \end{subfigure}~
     \hspace{0.15cm}
     \begin{subfigure}[b]{0.15\textwidth}
         \centering
         \includegraphics[height=1.6\textwidth]{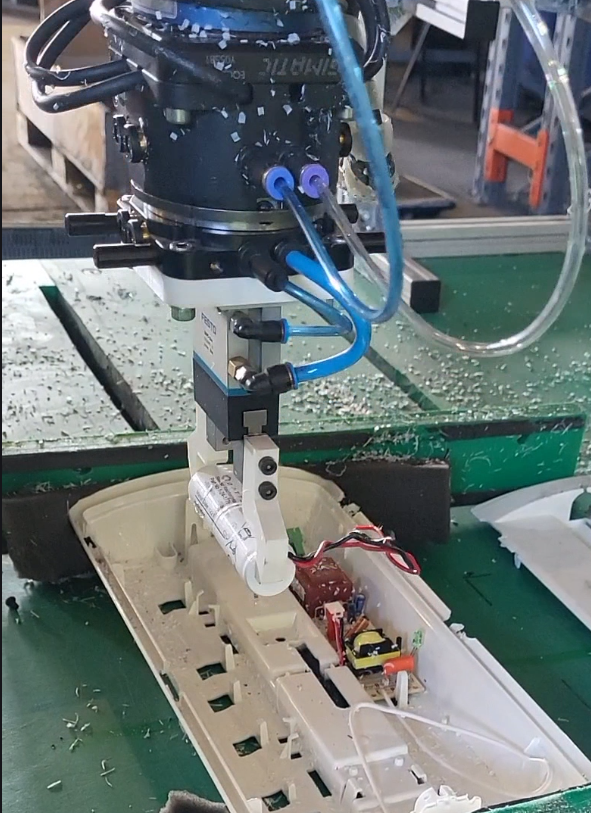}
         \caption{Battery (B).}
         \label{fig:battery}
     \end{subfigure}~
     \hspace{2cm}

\vspace{.2cm}
     \hspace{2.1cm}
     \begin{subfigure}[b]{0.15\textwidth}
         \centering
        \includegraphics[height=1.6\textwidth,origin=c, trim={0 0 0  5cm},clip]{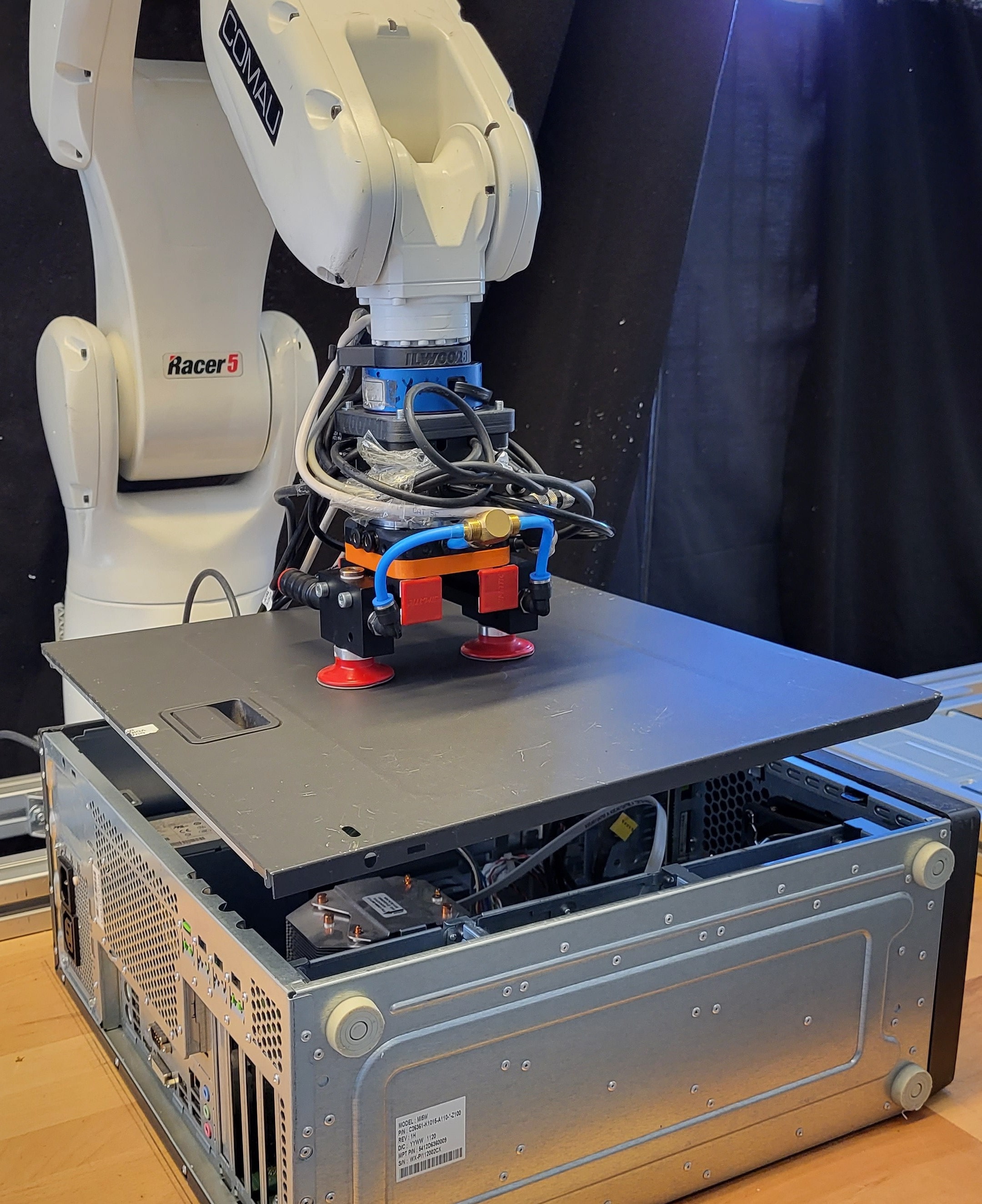}
         \caption{Top cover (C).}
         \label{fig:topcovPCT}
     \end{subfigure}
     \hspace{0.47cm}
     \begin{subfigure}[b]{0.15\textwidth}
         \centering
         \includegraphics[height=1.6\textwidth, trim={0 0 0  2cm},clip]{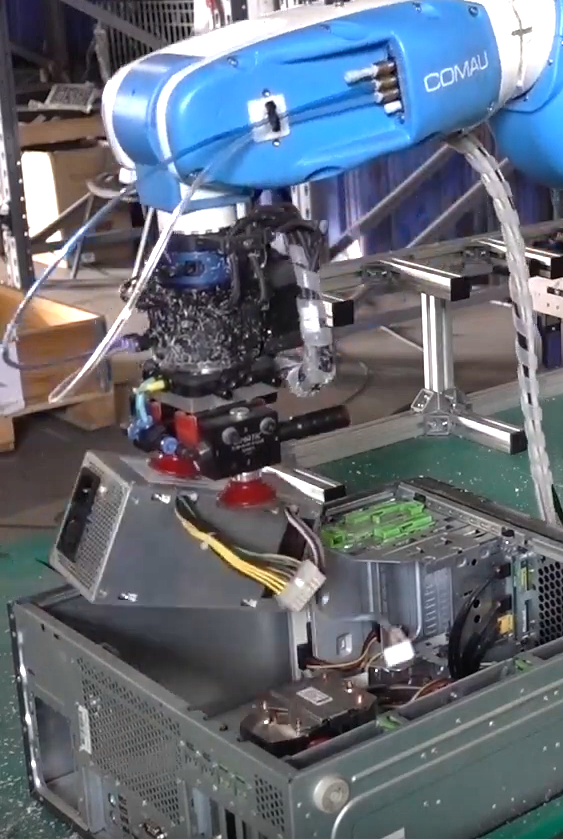}
         \caption{PSU (C).}
         \label{fig:PSU}
     \end{subfigure}
     \hspace{0.3cm}
     \begin{subfigure}[b]{0.15\textwidth}
         \centering
         \includegraphics[height=1.6\textwidth,origin=c,trim={0 0 0 12cm},clip]{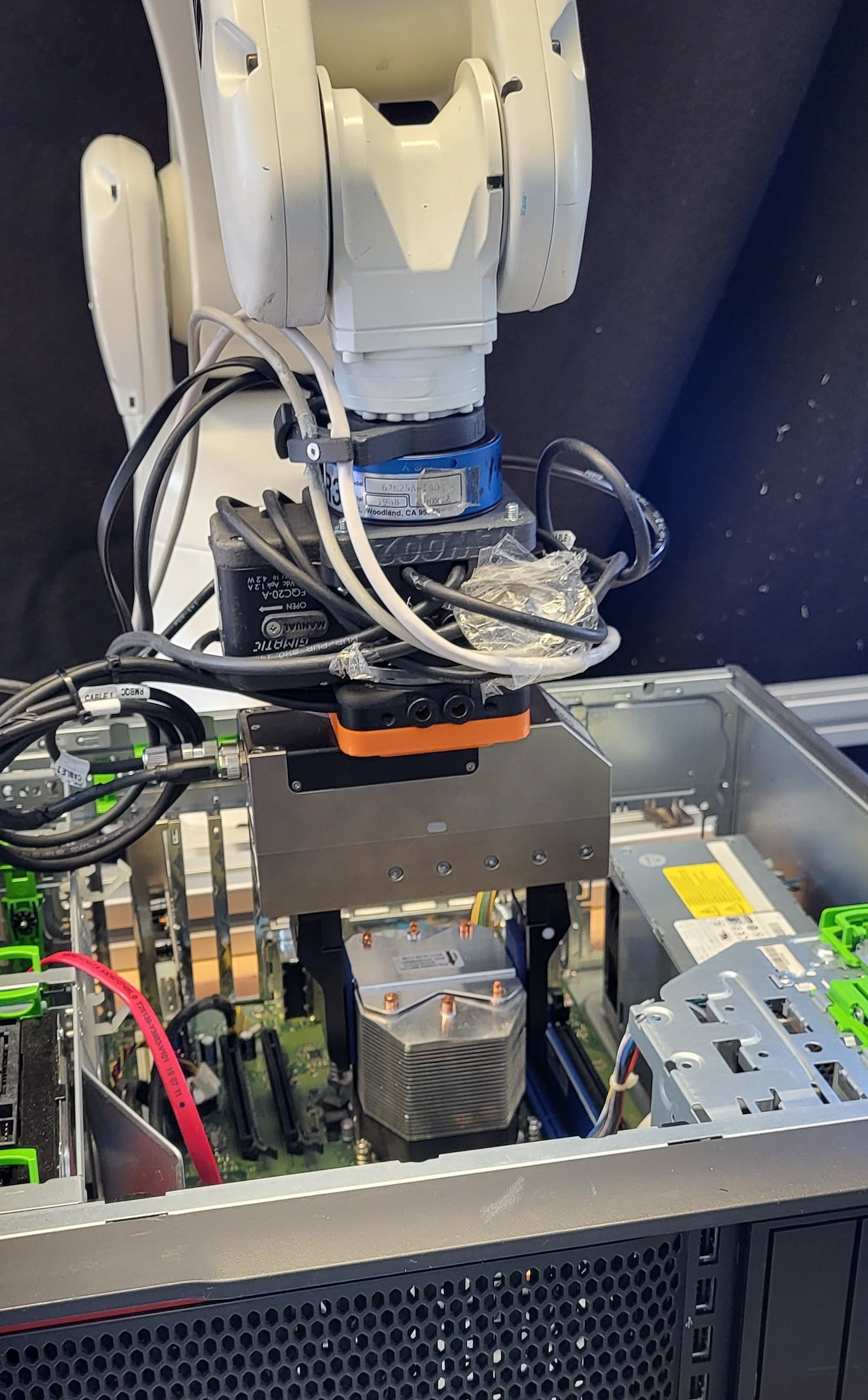}
         \caption{Cooler (A).}
         \label{fig:CoolerA}
     \end{subfigure}
    \hspace{0.3cm}
     \begin{subfigure}[b]{0.15\textwidth}
         \centering
         \includegraphics[height=1.6\textwidth]{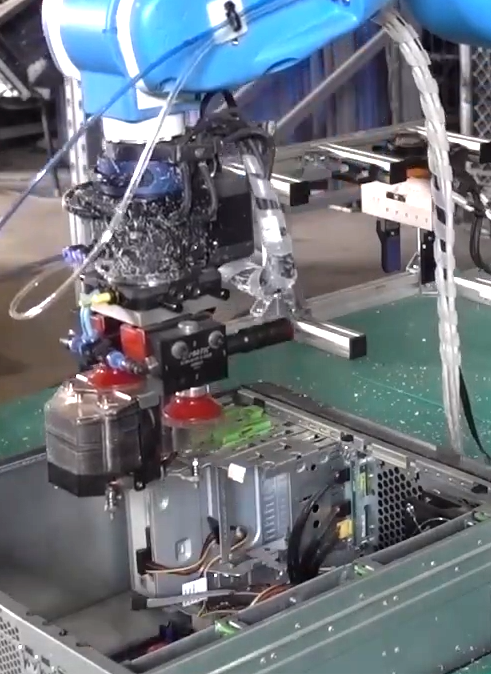}
         \caption{Cooler (C).}
         \label{fig:CoolerC}
     \end{subfigure}
     \hspace{2 cm}
     \vspace{-0.1cm}
    \caption{Two examples of the grasping skills used for the disassembly of the \ac{EL} (Subfig. a - d) and the \ac{PCT} (Subfig. e - h). The used grasping strategy is indicated in brackets. Subfig. a, c, e, g show the grasping in the lab environment, while Subfig. b, d, f, h are set at the industrial recycling company.}
        \label{fig:Examples}
        \vspace{-0.3cm}
\end{figure*}

\vspace{-0.1cm}
We validated our proposed gripping strategies within the EU project "HR-Recycler", which aimed to develop an innovative solution for the pre-processing of \ac{WEEE} indoors by replacing the hazardous and time-consuming manual tasks with automatic robotic procedures. As a first step, the required capabilities for dismantling four types of \ac{WEEE} (\ac{MW}, \ac{PCT}, \ac{EL}, \ac{FPD}) were determined by setting up the system in the laboratory. Here, it was found that simple grasping skills with small deviations in the target pose almost always failed. To this end, the introduced controllers and grasping algorithms were developed. The objects grasped and the respective applicable grasping strategy are described in Tab.~\ref{tab:objects}. In the lab environment, where solely the grasping skills were tested, assumptions had to be made about the accuracy of the visual input data~\eqref{eq: xnd}. A very high success rate was ultimately achieved, where all grasping skills succeeded in the vast majority of attempts. In addition, the grasping capabilities were also tested in a full-scale industrial recycling plant in Athens, Greece, where they were integrated into a higher-level process with a task planner and a vision unit, using real-world data for the capabilities that could be generated with state-of-the-art recognition algorithms. Compared to an approach without the hybrid control and the tactile methods we developed for the grippers, the success rate was significantly improved. However, since in this case, the result is highly dependent on the other two parties, no concrete success rate numbers can be given for the gripping skills alone. Nevertheless, it is clear that our approach now enables object grasping even if the initial vision data regarding the grasping point, orientation, and object dimensions are wrong by several millimeters (even centimeters in some applications). Fig.~\ref{fig:Examples} shows an overview of examples of the gripping strategies used for the \ac{EL} and \ac{PCT}. Half of the images were taken in the lab and the other half in the plant.

\vspace{-0.1cm}
\section{Conclusion}
\vspace{-0.1cm}
In this research paper, we introduced three grippers and presented three distinct grasping strategies suitable for uncertain environments. The key idea was to show force-feedback methods that help render object grasping more robust against uncertainty in object shape and pose. Our study focused on disassembling four different types of \ac{WEEE} devices where vision data is unreliable, and tactile data is essential. Through experiments conducted in both lab and factory settings, we demonstrated the cost-effectiveness of our approach and showed how object uncertainty can be overcome in a flexible robotic disassembly cell. The result was a significant increase in grasping success rates. We concluded that each gripper and its algorithm can effectively handle various grasping tasks with their unique strengths.
The presented grasping algorithms are not limited to our specific application and can be used in other scenarios that require grasping in uncertain environments with similar requirements. The algorithms can be tailored to meet the specific requirements of different grippers, such as adjusting for opening width and weight capacity. However, successful recycling cells need additional tools to perform tasks such as robust unscrewing and fixture loosening, which are the subject of future works. Additionally, the grasping of objects with more complex shapes and the accomodation of soft grippers remains a future research area.
\bibliography{literature}
\bibliographystyle{unsrt}

\end{document}

%% file: preamble.tex
\usepackage{cite}
\usepackage{amsmath,amssymb,amsfonts}
\usepackage{todonotes}
\usepackage{algorithmic}
\usepackage{graphicx}
\usepackage{textcomp}
\usepackage{caption}
\usepackage{subcaption}
\usepackage{multirow}
\usepackage{tikz}
\usepackage{adjustbox}
\usepackage{array}

\usepackage{color, colortbl}

\usetikzlibrary{shapes.geometric}
\usetikzlibrary{arrows.meta,arrows}

\def\BibTeX{{\rm B\kern-.05em{\sc i\kern-.025em b}\kern-.08em
    T\kern-.1667em\lower.7ex\hbox{E}\kern-.125emX}}
    
\newcommand{\greencheck}{}%
\DeclareRobustCommand{\greencheck}{%
  \tikz\fill[scale=0.4, color=black!60!green]
  (0,.35) -- (.25,0) -- (1,.7) -- (.25,.15) -- cycle;%
}    

\newcommand{\orangecheck}{}%
\DeclareRobustCommand{\orangecheck}{%
  \tikz\fill[scale=0.4, color=orange]
  (0,.35) -- (.25,0) -- (1,.7) -- (.25,.15) -- cycle;%
}

\newcommand{\redcross}{}%
\DeclareRobustCommand{\redcross}{%
  \tikz[scale=0.23, color=red] {
    \draw[line width=0.7,line cap=round] (0,0) to [bend left=6] (1,1);
    \draw[line width=0.7,line cap=round] (0.2,0.95) to [bend right=3] (0.8,0.05);
}
}
\newcolumntype{R}[2]{%
    >{\adjustbox{angle=#1,lap=\width-(#2)}\bgroup}%
    l%
    <{\egroup}%
}

\input{sample.tikzstyles}
\input{sample.tikzdefs}

\usepackage[nolist]{acronym}


%% file: sample.tikzstyles
\tikzstyle{red dot}=[fill=red, draw=red, shape=circle]
\tikzstyle{medium box}=[fill=white, draw=black, shape=rectangle, minimum width=0.75 cm, minimum height=1cm, rounded corners=0.2cm, line width=1 pt]
\tikzstyle{big box width}=[fill=white, draw=black, shape=rectangle, minimum width=15 cm, minimum height=1 cm, rounded corners=0.2 cm, line width=1 pt]

\tikzstyle{DoubleRed}=[<->, draw=red, arrows={Latex[length=5mm]-Latex[length=5mm]}, line width=1pt]
\tikzstyle{RightArrow}=[->, line width=1pt, arrows={-Latex[length=3mm]}]
\tikzstyle{DoubleArrow}=[<->, arrows={Latex[length=3mm]-Latex[length=3mm]}, line width=1 pt]
\tikzstyle{Endright}=[{-|}, line width=1 pt, arrows={-Rays[red]}]
\tikzstyle{dashed right}=[->, style=dashed, line width=1 pt, arrows={-Latex[length=3mm]}]

%% file: acronyms.tex
\begin{acronym}
    \acro{WEEE}{Waste Electrical and Electronic Equipment}
    \acro{EL}{Emergency Lamp}
    \acro{FPD}{Flat Panel Display}
    \acro{MW}{Microwave}
    \acro{PCT}{PC Tower}
    \acro{FT}{Force-Torque}
    \acro{CoP}{Center of Pressure}
    \acro{EE}{End-Effector}
    \acro{TCP}{Tool Center Point}
\end{acronym}